# Mining for Adverse Drug Events with Formal Concept Analysis

Alexander ESTACIO-MORENO[a,1], Yannick TOUSSAINT[a], Cédric BOUSQUET[b, c]
[a] *LORIA, Campus Scientifique. BP 239, 54506 Vandoeuvre. Nancy Cedex, France*
[b] *INSERM UMR_S 872, Eq 20, Faculté de Médecine - Paris 5, France*
[c] *University of Saint Etienne, Department of Public Health and Medical Informatics, St-Etienne, France.*

**Abstract**: The pharmacovigilance databases consist of several case reports involving drugs and adverse events (AEs). Some methods are applied consistently to highlight all signals, i.e. all statistically significant associations between a drug and an AE. These methods are appropriate for verification of more complex relationships involving one or several drug(s) and AE(s) (e.g; syndromes or interactions) but do not address the identification of them. We propose a method for the extraction of these relationships based on Formal Concept Analysis (FCA) associated with disproportionality measures. This method identifies all sets of drugs and AEs which are potential signals, syndromes or interactions. Compared to a previous experience of disproportionality analysis without FCA, the addition of FCA was more efficient for identifying false positives related to concomitant drugs.

## 1. Introduction

Detecting unexpected relationships between drugs and adverse events (AEs) from a pharmacovigilance database is a real challenge. Automated signal detection consists in application of data mining algorithms to help identification of signals, i.e. all statistically significant {drug, AE} couples, which are numerous and should be reduced to the most plausible [1]. Indeed, these potential signals extracted from a database are only assumptions; they should lead to more complex and expensive pharmacological studies. Therefore, it's important not to ignore any relationships but avoid an excess of studying situations.

Several types of relationships can be identified from the database: (i) {drug, AE} couples that data mining algorithms have mainly focused on; (ii) Relationships between, two drugs and an AE which are potential drug interactions, (iii) Relationships between several AEs and a drug which are potential syndromes, (iv) Relationships involving several drug(s) and AE(s) which are frequently observed in protocols. Among relationships between several drugs, cases where the drug association strengthens a signal compared to both single drugs may be classified as a potential interaction. Cases where there is no signal between one of the drugs and the AE are more difficult to interpret and may be related to concomitant drugs: the first drug is responsible for the AE but the second drug is only an innocent bystander which was associated to the first drug by chance or due to a frequent therapeutic association.

---
[1] Corresponding Author

Automated methods for detecting drugs-AEs relationships are based on calculation of disproportionality measures. The statistical approach includes the Proportional Reporting Ratio (PRR) [2], $\chi^2$ often coupled with the PRR, and the Reporting Odds Ratio [1]. The Bayesian approach consists of the Multi-item Gamma Poisson Shrinker algorithm [3] and the Bayesian Confidence Propagation Neural Network [4][2]. Recently, van Puijenbroek et al have used these methods in the detection of drug interactions [5] using a model based on logistic regression. These methods are appropriate for verification of more complex relationships involving several drugs and/or AEs (e.g; syndromes or interactions) but do not address the identification of them. A syndrome detected by case review involving Terbinafine and three AEs (arthralgia, urticaria and fever) was verified using this model [6]. A couple of articles addressed verification of already known drug interactions based on pharmacological knowledge. No systematic research and verification of relationships involving several drugs and/or syndromes was described in the literature with data mining algorithms.

Our new approach detects all kinds of drugs-AEs relationships thanks to an exhaustive exploration of a pharmacovigilance database. It helps identifying situations where drugs are prescribed together whereas a previous method [7] generated several false positives due to the innocent bystanders. It is based on Formal Concept Analysis (FCA) [8] combined with disproportionality measures. This article introduces briefly FCA and its use in pharmacovigilance, and then presents the results of an application of this method to an extract of the French pharmacovigilance database. We finally conclude with a discussion of the results.

## 2. Materials and Methods

*2.1. Fundamentals of Formal Concept Analysis (FCA)*

The central notion in Formal Concept Analysis [8] is a hierarchical structure called Concept Lattice where objects are classified following the attributes they own. The set of objects ($O$) and attributes ($A$) together with their relation to each other ($I \subseteq O \times A$) is called a *formal context* $K = (O; A; I)$. Table in figure 1(a) defines a formal context for adverse drug event exploration, where $O$ is a set of patients and $A$ is the set of attributes: sex (male and female), drugs that patients took (D1, D2, D3, D4, D5) and adverse events (AE1, AE2). Two derivation operators, both denoted by ' link objects and attributes. Let $X \subseteq O$ and $Y \subseteq A$ then $X' = \{a \in A / \forall o \in X, oIa\}$ and $Y' = \{o \in O / \forall a \in Y, oIa\}$. Two compound operators both noted " are defined by the composition of the two ' operators (i.e. " = ' o '). They are *closure operators* over $2^O$ (": $O \rightarrow O$) and $2^A$ (": $A \rightarrow A$). A pair of '-connected sets is called *formal concept*: $(X,Y) \in O \times A$ is a formal concept iff $X' = Y$ and $Y' = X$. $X$ and $Y$ are respectively the *extent* and the *intent* of the concept. The set $C_K$ of all concepts of the context $K$ is partially ordered by inclusion $\leq_K$ on $2^O$: $(X_1, Y_1) \leq_k (X_2, Y_2) \Leftrightarrow X_1 \subseteq X_2$. $L_k = \langle C_k, \leq_k \rangle$ is a *concept lattice*.

---

[2] See [1] for a comparison of these methods

In the context given figure 1(a), $\{P4, P5\}' = \{F\}$ but, following the previous definition, the formal concept built by closure is $(\{P4, P5, P6, P7, P8\}, \{F\})$. The lattice $L_k = \langle C_k, \leq_k \rangle$ has one *top* concept which extent groups all the objects of the context and, by duality, one *bottom* concept which intent groups all the attributes of the context. Objects in the extension of a concept are inherited from the bottom to the top of the lattice. In a dual way, attributes are inherited from top to bottom. Figure 1(b) is the concept lattice of the context (a). The concept $(\{P1, P2, P3, P4\}, \{AE1, D1\})$ is named $c_3$ in the lattice.

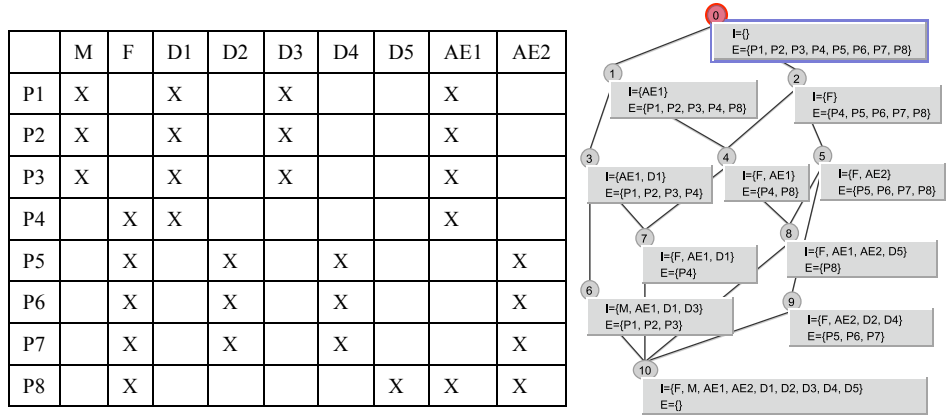

|    | M | F | D1 | D2 | D3 | D4 | D5 | AE1 | AE2 |
|----|---|---|----|----|----|----|----|-----|-----|
| P1 | X |   | X  |    | X  |    |    | X   |     |
| P2 | X |   | X  |    | X  |    |    | X   |     |
| P3 | X |   | X  |    | X  |    |    | X   |     |
| P4 |   | X | X  |    |    |    |    | X   |     |
| P5 |   | X |    | X  |    | X  |    |     | X   |
| P6 |   | X |    | X  | X  |    |    |     | X   |
| P7 |   | X |    | X  | X  |    |    |     | X   |
| P8 |   | X |    |    |    | X  | X  | X   |     |

**Figure 1**: **(a)** Context describing patients. **(b)** The corresponding concept lattice

*2.2. FCA and Adverse Drug Event Exploration*

The intent of some formal concepts shows that some patients took a same set of drugs and had the same set of AEs. In $c_3$, four patients $\{P1, P2, P3, P4\}$ took the drug $D1$ and have the adverse event $AE1$. We also observe in the context that $P3$, among others, also took $D3$ but $D1$ is the only drug that the four patient $\{P1, P2, P3, P4\}$ took. Thanks to the closure operators used to build the concepts, the lattice gives all the drugs-AEs relationships that exist in the data and no more. Among all the concepts, we distinguish potential **signals**, when the intent includes only one drug and one AE (see $c_3$ and $c_7$); potential **drug interactions**, when the intent includes two drugs and one AE (see $c_6$ and $c_9$); and potential **syndrome**, when the intent includes one drug and several AEs (see $c_8$).

Compared to extraction of {drug, AE } couples without FCA, the lattice – with its formal concepts built using the closure operators – reduces the number of situations drugs-AEs to study. In figure 1(b), patients $\{P1, P2, P3\}$ share the attributes $\{AE1, D1, D3\}$. While disproportionality measures would propose two potential signals to study ($\{AE1, D1\}$ and $\{AE1, D3\}$), the lattice keeps only one ($\{AE1, D1\}$) as $\{AE1, D3\}$ is not shared by any group of patients. Of course, thresholds used in disproportionality measures are not meaningful on such a small example.

All formal concepts are not relevant for adverse drug event exploration. Either they do not describe a drug-AEs relationship or the described relationship is not statistically significant. To keep statistically significant concepts when the intent includes drugs and AEs, we combine the lattice with criteria used by the British Medicines and Healthcare products Regulatory Agency (MHRA): $\chi^2 > 4$, PRR > 2 and the number of patients (the support) is greater or equal to 3 (see [2] for $\chi^2$ and PRR calculation). The *a,b,c* and *d* parameters for $\chi^2$ and PRR calculation are calculated from the extent of concepts in the lattice. We tested this method to a subset of 3249 cases from the French database on AEs. Drugs where encoded as active ingredients and AEs where encoded following the WHO-ART terminology. In that way, our database contains 527 drugs and 639 AEs. First, the lattice has been built using the Galicia software and then, for any concept whose support is higher than 3, we calculated both PRR and $\chi^2$ values.

## 3. Results

The whole lattice has 13,178 concepts. For 842 of them, the intent includes at least one drug and one AE. Relationships are investigated using MHRA criteria in a horizontal strip of the lattice (see black circles in figure 2b). This remaining strip is delimited in the upper part according to the meets, which are the first concepts of the lattice starting from the *top* including both drugs and AEs in their intent. Delimitation in the lower part is according to the support of concepts because we keep concepts including at least three patients in their extent.

| Concepts | Count |
|---|---|
| Total | 13,178 |
| $\{D_1, \ldots D_n; AE_1 \ldots AE_m\}$ | 842 |
| $\{D_1, \ldots D_n; AE_1 \ldots AE_m\}$ with PRR>2, $\chi^2$>4, Support ≥ 3 | 593 |

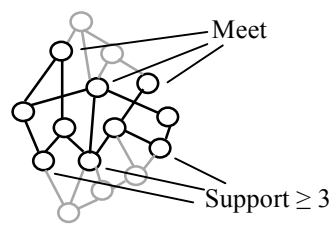

**Figure 2**: Concepts of the lattice before and after filtering with the MHRA criteria. **(a)** Results of our application on 3249 pharmacovigilance cases. **(b)** A graphical representation of conserved concepts (black circles) in a fictitious lattice.

**Table 1.** Drug-AEs relationships detected with FCA after filtering with the MHRA criteria

| Relationship | Count | Type |
|---|---|---|
| $\{D_1; AE_1\}$ | 360 | Potential signal |
| $\{D_1, D_2; AE_1\}$ | 110 | Potential drug interaction |
| $\{D_1; AE_1 \ldots AE_m\}$ | 56 | Potential drug syndrome |
| $\{D_1, \ldots D_n; AE_1\}$ | 42 | Potential complex interaction |
| $\{D_1, \ldots D_n; AE_1 \ldots AE_m\}$ | 25 | Potential complex syndrome |
| Total | 593 | PRR>2, $\chi^2$>4, support ≥ 3 |

Table 1 shows five different kinds of drug-AEs detected relationship. 181 among the 360 concepts expressing a potential signal are not restricted to a given population (their intent do not include any patient properties, neither age nor sex) We evaluated a sample of 95 of them: 61 signals where already known, for example {Abciximab, Thrombopenia}; 34 were not described in the literature, for example {Lidocaine, Tachycardia} and need further investigations. Among the 56 potential syndromes, 53 consisted of two AEs and three consisted of three AEs.

**Table 2.** Classification of potential drug interactions

| Drug association | Count | Example |
| --- | --- | --- |
| Therapeutic associations | 57 | Fluorouracil + oxaliplatine |
| Successive prescriptions | 8 | Heparin + enoxaparin |
| Medicinal product | 22 | Amoxicillin + clavulanic acid |
| Concomitant drugs | 23 | Heparin + sucralfate |
| Total | 110 | |

The lattice helps to identify 110 pairs of drugs that might be classified as drug interactions. Among 57 pairs related to drugs commonly prescribed in therapeutic associations, 34 were dedicated to anti-HIV prescriptions. In 8 cases the associated drugs were likely to be prescribed successively rather than at the same time for example heparin was prescribed first then replaced by enoxaparin.

**4. Discussion and conclusion**

We evaluated in this article the properties of the concept lattice for generating potential relationships between one or several drugs and AEs. Sample size was sufficient for generating several signals. Syndromes mainly consisted of two AEs. No database on syndromes is available to verify their plausibility but we were able to detect known patterns such as hypersensitivity to abacavir. FCA also highlighted more complex relationships that are more difficult to interpret because they are associations of more than two drugs and AEs. This could help the generation of signals related to a whole protocol instead of single drugs. Due to sample size few drug interactions were detected and the interest of the lattice was more focused on the identification of false positives and concomitant drugs.

In a previous experience [7], disproportionality measures extracted 523 statistically significant {drug, AE} couples with many false positives due to concomitant prescription. The new approach generates only 360 signals. Indeed many signals generated with the previous approach are now clustered in concept nodes of the lattice that present higher order relations, i.e. several drugs and/or AEs. FCA helps to reduce:

1. The number of signals to review. For example the medicinal product consisting of tiemonium + colchicine + opium generated a single signal for the whole association whereas the previous method generated three different signals related to each single active ingredient.
2. The number of false positives related to several active ingredients in the same medicinal products. In previous method a signal of diarrhoea was generated with opium whereas this active ingredient is added in small proportions to colchicine in order to reduce the diarrhoea episodes. Compound active ingredients are clustered by the current method and it is easier to interpret diarrhoea as related to colchicine.
3. The number of false positives related to concomitant drugs. For example sucralfate was no longer associated to thrombopenia while all patients taking sucralfate were also prescribed heparin.

We showed how FCA may be combined with disproportionnality measures for the identification of three kinds of relationships of great interest in pharmacovigilance: signals, drug interactions and syndromes. Building formal concepts by closure with FCA ensures that resulting relationships (signals, interactions and syndromes) are relevant relationships present in data. Thus in the case of signals, the count of associations the expert has to evaluate is reduced from the number of signals that would be generated by a classical signal detection method.

The method we presented performs exhaustive search of the three kinds of existing relationships in data. This suggests new relationships for evaluation by experts, especially in the case of interactions and syndromes, present in data with no a priori of the domain.

## Acknowledgments

We acknowledge Dr Agnès Lillo-Le Louët, director of the Pharmacovigilance Centre of the Georges Pompidou hospital for providing an extract of the French Pharmacovigilance database and Dr Pascal Auriche from the *Agence française de sécurité sanitaire des produits de santé* for extracting the data. Alexander Estacio Moreno was supported by a grant from the French « Agence Nationale de la Recherche » (ACI Masse de données).